# Human Assisted Artificial Intelligence Based Technique to Create Natural Features for OpenStreetMap

| Piyush Yadav | Dipto Sarkar | Shailesh Deshpande | Edward Curry |
|---|---|---|---|
| NUI Galway Ireland | Carleton University Canada | TCS Research India | NUI Galway Ireland |

## Introduction

OpenStreetMap (OSM) is arguably the largest crowdsourced geographic databases with more than one million contributors [1]. The ongoing contributions make OSM an ever-evolving spatial dataset with the improving quantity, quality, and coverage of data across all types of map features. While OSM has stood up to various tests of data quality and accuracy with regards to a variety of manmade features (e.g. building, roads), data completeness of high-level natural feature classes like vegetation (grass, forest), impervious surface (industrial regions) and water (lakes, river, delta) is relatively sparse. This is partly because OSM data creation tools (e.g. iD) provide an intuitive interface to digitize smaller features, such as building and road segments, but is cumbersome for natural features (e.g. delta, scree, geological regions) which are quite complex and are spread over larger regions.

Recently, Facebook presented RapiD editor[1] which uses Artificial Intelligence (AI) techniques to generate features for faster map editing. RapiD autodetects roads features from high-resolution satellite imagery which are then curated by users to create the final dataset. The RapiD program has certain limitations: for example, discovering linear features like road is easier as compared to natural elements (such as delta and forest) which are uneven [2]. Secondly, most of the high-resolution imagery is still not freely available which contradicts the preamble of the open-source community. In this work, we propose an AI-based technique using freely available satellite images like Landsat and Sentinel to create natural features over OSM in congruence with human editors acting as initiators and validators. The method is based on Interactive Machine Learning [3] technique where human inputs are coupled with the machine to solve complex problems efficiently as compare to pure autonomous process. We use a bottom-up approach where a machine learning (ML) pipeline in loop with editors is used to extract classes using spectral signatures of images and later convert them to editable features to create natural features.

## Proposed Approach

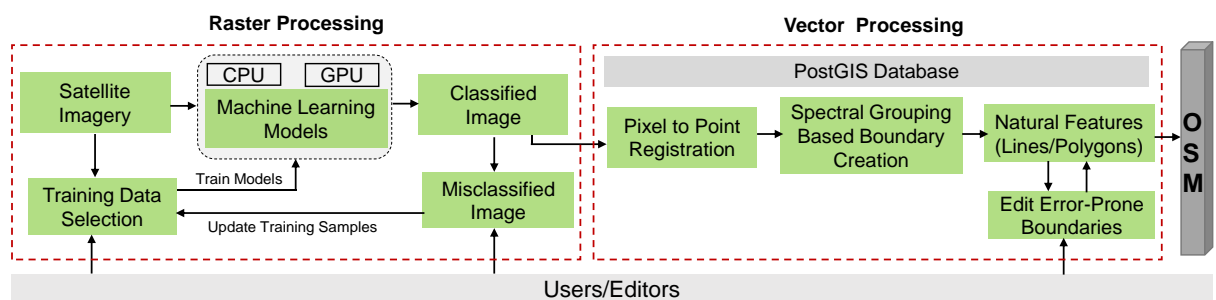

**Figure 1.** High-level architecture for the proposed approach

As per Fig.1, our proposed method is divided into two-part:

---
[1] https://mapwith.ai/

**Low-level Raster Processing:** To start the process, the user labels training samples belonging to different classes by selecting pixels from the imagery. It is assumed that the user has some knowledge about the natural features in the area. The number of macro (e.g. water, vegetation) or micro (e.g. lake and grass) classes can be varied depending on the user's expertise of the region. The ML models are then trained using the training data. The proposed system consists of a suite of different ML models like Support Vector Machine (SVM), Nearest Neighbour (NN) and Convolutional Neural Networks (CNN) running on GPU instances for efficient processing. The models predict the class of every pixel and classify the image. The misclassified regions can again be selected by the user to improve model accuracy by adding new training samples iteratively. Finally, the image registration is performed over a classified image so that each pixel spatial location is known.

**Top-level vector processing:** At top-level image primitives and complex objects are created using the proposed grouping method.

*Image Primitives:* A spatial entity can be represented using geometry-based features like points, lines, and polygons. The point is a zero-dimensional image primitive located in cartesian coordinate while a line represents a one-dimensional object defined by a list of points. Similarly, a polygon is a two-dimensional bounded region created from a list of lines.

*Organization of Image Primitives:* On a high-level abstraction, an image can be represented as a collection of points representing each pixel in the image. The classified image pixels are converted to points where each point represents a spatial location with a known class (e.g. water). A grouping-based algorithm is devised based on class similarity and vicinity to combine points to line and lines to polygons. The grouping occurs in the following order: 1) if adjacent points have same class then group them, 2) the first level grouping is then converted to boundary edges (line) and closed polygons.

The intermediate image primitives are stored in a PostGIS database where elements can be fetched using the query. The editors can manually refine the error-prone automated boundaries which can then be exported to OSM after validation.

## Results

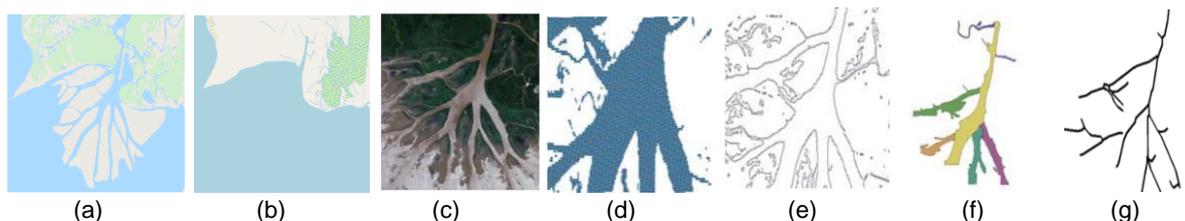

(a) (b) (c) (d) (e) (f) (g)

**Figure 2.** Wax Lake Delta (a) Google Map View (b) OSM view (c) Landsat 8 Image (d) Pont representation of water pixels (e) Waterbody Boundary (f) Polygon for different water streams (g) Waterbody medial skeleton

The proposed approach is implemented in Python 3. For vector processing, PostGIS database is used to store the intermediate results using pyscopg2 adapter. The area of study is Wax Lake River Delta in the United States to identify complex delta boundaries [4]. Fig 2 (a) and (b) shows the comparison between Google Maps and OSM of the study area, respectively, where OSM is missing the delta boundaries. The Landsat 8 image of the selected region (Fig. 2 (c)) is classified using SVM classifier where soil, water and vegetation pixels are classified based on their spectral signatures. The misclassified region training samples were again collected to improve classification accuracy. Fig. 2 (d) shows the water pixels as vector points which are later connected to form boundary (Fig.2 (e)), polygons (Fig. 2 (f)) and lines (Fig. 2 (g)) using class-based grouping algorithms. QGIS is used as a front-end editing tool for detected features as it consists of robust vector processing suites to edit the features. The final curated features are then can be exported to OSM from bottom to top levels, i.e. point and line and polygon.

## Conclusion

In this work, we propose a human assisted-ML framework to contribute natural features to OSM. This tool works in conjunction with the open-source GIS platform QGIS and provides an intuitive interface to create natural features spread over a large spatial extent. The contributions made by the users using this tool will help OSM move closer to being a complete spatial data repository. We plan to release this tool as a plugin for QGIS in the coming months.